\documentclass[11pt,a4paper]{article}
\usepackage[hyperref]{acl2020}
\usepackage{times}
\usepackage{latexsym}

\usepackage{microtype}

\usepackage{graphicx} 
\usepackage{float} 
\usepackage{subfigure} 
\usepackage{amsmath}
\usepackage{booktabs}
\usepackage{multirow}

\usepackage{xcolor}

\aclfinalcopy 
\def\aclpaperid{2517} 

\title{Improving Readability for Automatic Speech Recognition Transcription
}

\author{
  {Junwei Liao}$^{1}$\footnotemark[1] \quad {Sefik Emre Eskimez}$^2$ \quad {Liyang Lu}$^2$ \quad {Yu Shi}$^2$\footnotemark[2] \quad {Ming Gong}$^3$ \\
  \textbf{Linjun Shou}\textsuperscript{3} \quad \textbf{Hong Qu}\textsuperscript{1} \quad \textbf{Michael Zeng}\textsuperscript{2}\\
  $^1${University of Electronic Science and Technology of China} \\
  $^2${Microsoft Speech and Dialogue Research Group} \\
  $^3${Microsoft STCA NLP Group} \\
  \texttt{junwei.liao@outlook.com} \\
  \texttt{\{seeskime, liyang.lu, yushi, migon, lisho, nzeng\}@microsoft.com} \\
  \texttt{hongqu@uestc.edu.cn} \\
}

\date{}

\begin{document}
\maketitle

\renewcommand{\thefootnote}{\fnsymbol{footnote}} 
\footnotetext[1]{Work is done during internship at Microsoft.} 
\footnotetext[2]{Corresponding author.} 
\renewcommand{\thefootnote}{\arabic{footnote}} 

\begin{abstract}



Modern Automatic Speech Recognition (ASR) systems can achieve high performance in terms of recognition accuracy. However, a perfectly \textit{accurate} transcript still can be challenging to read due to grammatical errors, disfluency, and other errata common in spoken communication. Many downstream tasks and human readers rely on the output of the ASR system; therefore, errors introduced by the speaker and ASR system alike will be propagated to the next task in the pipeline. In this work, we propose a novel NLP task called ASR post-processing for readability (APR) that aims to transform the noisy ASR output into a readable text for humans and downstream tasks while maintaining the semantic meaning of the speaker. In addition, we describe a method to address the lack of task-specific data by synthesizing examples for the APR task using the datasets collected for Grammatical Error Correction (GEC) followed by text-to-speech (TTS) and ASR. 

Furthermore, we propose metrics borrowed from similar tasks to evaluate performance on the APR task. We compare fine-tuned models based on several open-sourced and adapted pre-trained models with the traditional pipeline method. Our results suggest that fine-tuned models improve the performance on the APR task significantly, hinting at the potential benefits of using APR systems. We hope that the read, understand, and rewrite approach of our work can serve as a basis that many NLP tasks and human readers can benefit from. 

\end{abstract}

\section{Introduction}

With the rapid development of speech-to-text technologies, ASR systems have achieved high recognition accuracy, even beating the performance of professional human transcribers on conversational telephone speech in terms of Word Error Rate (WER) \citep{xiong2018microsoft}.

Automatic speech recognition systems bring convenience to users in many scenarios. However, colloquial speech is fraught with syntactic and grammatical errors, disfluency, informal words, and other noises that make it difficult to understand. While ASR systems do a great job in recognizing which words are said, its verbatim transcription creates many problems for modern applications that must comprehend the meaning and intent of what is said. Applications such as automatic subtitle generation and meeting minutes generation require automatic speech transcription that is highly readable for humans, while machine translation, dialogue systems, voice search, voice question answering, and many other applications require highly readable transcriptions to generate the best machine response. The existence of the defects in speech transcription will significantly harm the experience of the application users if the system cannot handle them well.

Inspired by the latest progress in natural language generation (NLG), grammatical error correction (GEC), machine translation, and transfer learning, we explore the idea of ``understanding then rewriting" as a new ASR post-processing concept to provide 
conversion from raw ASR transcripts to error-free and highly readable text.

We propose ASR post-processing for readability (APR), which aims to transform the ASR output into a readable text for humans and downstream NLP tasks. Readability in this context refers to having proper segmentation, capitalization, fluency, and grammar, as well as properly formatted dates, times, and other numerical entities. Post-processing can be treated as a style transfer, converting informal speech to formal written language.

Due to the lack of relevant data, we constructed a dataset for the APR task using a GEC dataset as seed data. The GEC dataset is composed of pairs of grammatically incorrect sentences and corresponding sentences corrected by a human annotator. First, we used a text-to-speech (TTS) system to convert the ungrammatical sentences to speech. Then, we used an ASR system to transcribe the TTS output. Finally, we used the output of the ASR system and the original grammatical sentences to create the data pairs. By this means, we produced 1.1 million APR samples that are used for training and testing.

We investigated three mainstream Transformer-based sequence-to-sequence neural network architectures for the APR task. Specifically, we investigated MASS~\citep{song2019mass}, UniLM~\citep{dong2019unified} and RoBERTa~\citep{liu2019roberta}, which are pre-trained models used for NLG and/or NLU tasks. We also attempted to leverage the advantages of both RoBERTa and UniLM by adapting the RoBERTa pre-trained model towards generative using a modified UniLM training approach (RoBERTa-UniLM).

We used several metrics to evaluate the four fine-tuned models on our APR dataset: readability-aware WER (RA-WER), BLEU, MaxMatch (M$^2$), and GLEU. The results show that the fine-tuned models outperform the baseline method significantly in terms of readability. 

Our main contributions can be summarized as follows:
\begin{itemize}
\itemsep=-4pt
\item We propose a novel task: ASR post-processing for readability (APR). It aims to solve the shortcomings of the traditional post-processing concept/methods by jointly performing error correction and readability improvements in one step. 
\item We describe a method to construct a dataset for the APR task. 
\item We experiment using state-of-the-art pre-trained models on the proposed APR dataset and achieved significant improvement on all metrics.
\item We adapt RoBERTa as a generative model trained in the style of UniLM which shows its benefits on some metrics such as M$^2$ and BLEU.
\end{itemize}

\section{Related Work}

\subsection{Automatic Speech Recognition (ASR)}

Traditional ASR systems take a pipelined approach,~\citep{paulik2008sentence,cho2012segmentation,cho2015punctuation,batista2008recovering,gravano2009restoring,vskodova2012discretion} relying on post-processing modules to improve the readability of the output text in two critical ways. First, a more robust language model is used to reduce word recognition errors via a second-pass rescoring of the output lattice or top recognition candidates. Then other sub-processes modify the sentence display format for readability using a series of steps, adding capitalization and punctuation, correcting simple grammatical errors, and formatting dates, times, and other numerical entities. We call these steps inverse text normalization (ITN). Originally, researchers mainly exploited handcrafted rules or statistical methods \citep{shugrina2010formatting,anantaram2016repairing,bohac2012post,liyanapathirana2016using,shivakumar2019learning,cucu2013statistical,bassil2012post} for post-processing.
Recently, \citet{guo2019spelling} trained an LSTM-based sequence-to-sequence model to correct spelling errors. \citet{hrinchuk2019correction} investigated the use of Transfomer-based architectures for the correction of SR output into grammatically and semantically correct forms. 

Traditional ASR post-processing methods offer improvements in readability; however, there are two important shortcomings.
(1) Since the whole process is divided into several sub-processes, the mistakes in the previous steps will accumulate. For example, in the sentence, ``Mary had a little lamb. It's fleece was white as snow.'', if in the punctuation step, a period `.' is added after the word ``had,'' the rule-based capitalization will capitalize the word`\emph{a}.' 
(2) The traditional methods tend to transcribe the speech verbatim while ignoring the readability of the output text. It cannot detect and correct disfluency in spontaneous speech transcripts. For example, in an utterance such as ``I want a flight ticket to Boston, uh, I mean to Denver on Friday'', the speaker means to communicate ``I want a flight ticket to Denver on Friday.'' The segment ``\emph{to Boston, uh, I mean}'' in the transcript is not useful for interpreting the intent of the sentence and hinders human readability and the performance of many downstream tasks. Traditional methods optimized for recognition accuracy will keep these words, increasing the cognitive load of the reader.

\subsection{Natural Language Processing (NLP)}
In NLP research, the most similar task to ours is automatic post-editing (APE) \citep{bojar2016findings}, which has been extensively studied by the machine translation (MT) community (e.g., \citealp{pal2016neural,pal2017neural,chatterjee2017multi,hokamp2017ensembling,tan2017neural}). These methods take input of the source language text, target language MT output, and target language post-editing (PE) for training. Based on our knowledge, there is no similar work in speech recognition field. 

Another similar task is the Grammatical Error Correction (GEC). GEC aims to correct different kinds of errors such as spelling, punctuation, grammatical, and word choice errors \citep{ge2018reaching,zhang2019sequence,napoles2019enabling,napoles2017jfleg,grundkiewicz2019neural,choe2019neural}. The difference between our task and GEC is that the latter aims to correct written language, while our task aims to correct spoken language that contains noise introduced by ASR errors as well that introduced by the discrepancy between spoken and written formats of natural language.
Due to the similarity between GEC and APR, we borrow some ideas from GEC and use GEC corpora as our seed corpus to synthesize our dataset and use GEC metrics, namely MaxMatch and GLEU, to evaluate APR performance. 

\subsection{Unsupervised Learning}

Pre-training approaches \citep{dai2015semi,mccann2017learned,howard2018universal} have drawn much attention recently, especially those that employ the Transformer~\citep{Vaswani2017AttentionIA} architecture. 
The most successful approaches are variants of masked language models, which are denoising autoencoders trained to reconstruct text where a random subset of the words has been masked out.
Among them, BERT~\citep{devlin2018bert} and RoBERTa~\citep{liu2019roberta} are single-stack Transformer encoders; GPT(-2)~\citep{radford2018improving,radford2019language} and XLNET\citep{yang2019xlnet} are single-stack Transformer decoders; UniLM~\citep{dong2019unified} is a single-stack Transformer serving both encoder and decoder roles; and MASS~\citep{song2019mass}, BART~\citep{lewis2019bart} and T5~\citep{raffel2019t5} are standard Tranformer-based neural machine translation architecture.
We use RoBERTa, UniLM, and MASS as our base architectures and use their pre-trained models for the APR task.

\begin{table}
\centering
\begin{tabular}{ll}
\toprule[1pt]
\textbf{Input} & \textit{She see Tom is catched by policeman} \\
 & \textit{in park at last night.} \\ 
\textbf{Output} & \textit{She saw Tom caught by a policeman} \\
 & \textit{in the park last night.} \\
\bottomrule[1pt]
\end{tabular}
\caption{\label{gec-example} A GEC data sample is shown. The input is a sentence with some grammatical errors. The output is a grammatically correct sentence.}
\end{table}

\section{Proposed Method}

\begin{figure*}[htbp] 
\centering 
\includegraphics[width=1\textwidth]{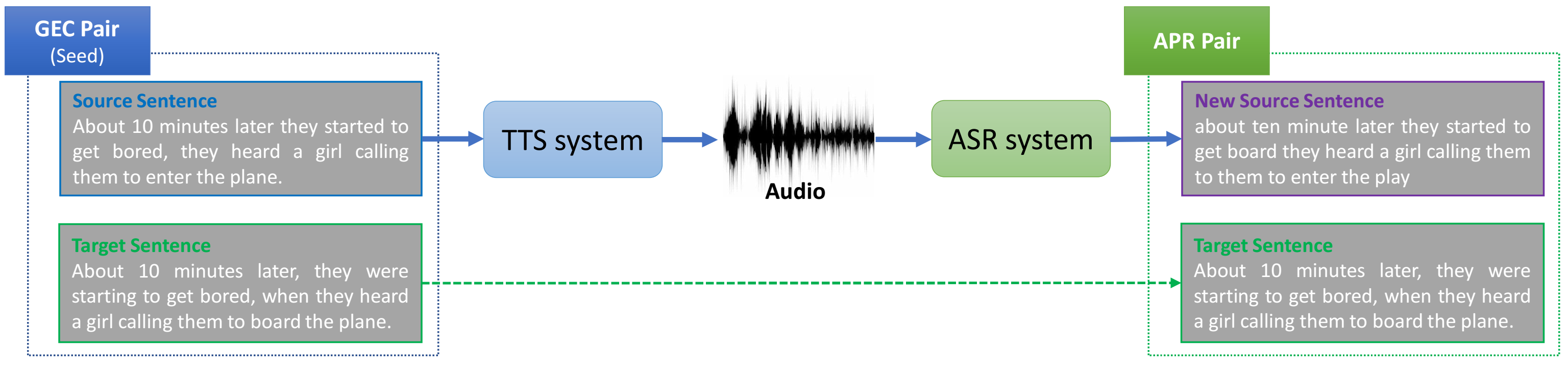} 
\caption{The process of data synthesis is shown. The left sentence is from the GEC dataset. The right sentence pair is an APR instance. The source sentence of GEC is processed by the TTS and ASR systems, and the APR sentence pair is obtained. The target sentence of GEC remains unchanged and is used as the target sentence for the APR.} 
\label{fig-data} 
\end{figure*}

\subsection{Dataset}

There exist a large amount of data that have been labeled for speech recognition. However, these data have two issues: first, they label the exact words that were spoken, including all disfluency. This is essential for HMM and hybrid acoustic model training but could hinder readability. Second, no capitalization and punctuation is present because spontaneous speech does not follow normal grammatical conventions. Similarly, entities, especially numerical entities, appear different in spoken language than when they appear in written form.


Due to these restrictions, we synthesize our data, simulating ASR errors by feeding sentences from a grammatical error correction (GEC) dataset into a text-to-speech (TTS) system and then transcribing this with an ASR model.

The GEC data samples contain grammatically correct and incorrect sentence pairs. A human corrects the grammatically incorrect sentence to obtain the target grammatically correct sentence. An example sentence pair from the seed corpus is shown in Table~\ref{gec-example}. Inspired by the GEC task, we simulated ASR errors using the GEC source sentences to obtain sentence pairs of which source sentences contain both grammatical errors and ASR errors. 

In the next section, we detail how we simulated the APR data. We discuss the simulated data statistics in \ref{sses:data_statistic}.

\subsubsection{Dataset Synthesis}
\label{ssec:data_synthesis}
We fed the grammatically incorrect sentences from the seed corpus into a neural-TTS system, which produced the audio files simulating human speakers. We used 320 different speaker voices for this simulation and split them into 220 for training, 50 for validation, and 50 for evaluation. Each sentence randomly selected one speaker, and all speakers have the same number of input sentences \cite{deng2018modeling}. Then these audio files are fed into the ASR system  that outputs the corresponding transcript. The resulting text contains both the grammatical errors found initially in the GEC dataset and the TTS+ASR pipeline errors. We used original corrected sentences as our target. The whole process is illustrated in Figure~\ref{fig-data}.

In addition to the mentioned simulation method, we tried using the top-k best output of the ASR system to augment our dataset ten-fold. However, we found that the augmented dataset is not beneficial, due to the lack of diversity in the resulting sentences, which often differ only in some characters (\ref{para:data-selection}).

\subsubsection{Dataset Statistic}
\label{sses:data_statistic}
Table~\ref{table-stat} shows dataset statistic of our data.

\begin{table}
\centering
\resizebox{\columnwidth}{!}{%
\begin{tabular}{c|c|l|r|r}
\toprule[1pt] 
\multicolumn{2}{c|}{\multirow{2}{*}{}} & \multicolumn{2}{c|}{\textbf{Seed corpus}} & \textbf{Synthetic data} \\
\cline{3-5}
\multicolumn{2}{c|}{} & \textbf{GEC dataset} & \textbf{sent pairs}  & \textbf{sent pairs} \\ 
\midrule[0.5pt]
\multicolumn{2}{c|}{\multirow{3}{*}{\textbf{training set}}} & FCE & 28,350 & \multirow{3}{*}{1,100,219}\\
\multicolumn{2}{c|}{} & W\&I+LOCNESS & 34,308 & \\
\multicolumn{2}{c|}{} & Lang-8 Corpus & 1,037,561 & \\
\hline
\multirow{2}{*}{CoNLL} & \textbf{dev} & CoNLL-2013  & 1,381 & 1,381 \\
\cline{2-5}
 & \textbf{test} & CoNLL-2014  & 1,312 & 1,312 \\
\hline
\multirow{2}{*}{JFLEG} & \textbf{dev} & JFLEG dev & 754 & 754  \\
\cline{2-5}
 & \textbf{test} & JFLEG test & 747 & 747\\
\bottomrule[1pt]
\end{tabular}
}
\caption{\label{table-stat} Dataset statistics are shown. We create synthetic data from the seed corpus using the synthesis process described in Section~\ref{ssec:data_synthesis}. Seed corpus FCE, W\&I+LOCNESS, and Lang-8 Corpus are used to synthesize the training data. Two datasets are used as evaluation data, which are evaluated by RA-WER and BLEU metrics. Specifically, following the GEC literature, the CoNLL dataset and JFLEG dataset are evaluated by MaxMatch and GLEU metrics, respectively.
}
\end{table}


We used the data from the datasets provided by restricted tracks of BEA 2019 shared task \citep{Bryant2019TheBS} as our seed corpora for training. Specifically, we collected data from FCE \citep{Yannakoudakis2011AND}, Lang-8 Corpus of Learner English \citep{Mizumoto2011MiningRL,Tajiri2012TenseAA}, and W\&I+LOCNESS \citep{Bryant2019TheBS,granger2014computer}, totaling to around 1.1 million training samples. 

Furthermore, we utilized CoNLL-2014 shared task dataset \citep{ng2014conll} and JFLEG \citep{napoles2017jfleg} test set as our evaluation seed corpora, to be aligned with the GEC literature \citep{ge2018reaching,zhang2019sequence,Kiyono2019AnES}. The CoNLL-2014 and JFLEG test sets contain 1,312 and 747 sentences, respectively. 

In order to be consistent with the standard evaluation metrics in the GEC literature, we used MaxMatch (M$^2$) $F_{0.5}$ \citep{Dahlmeier2012BetterEF} for CoNLL-2014 and used GLEU \citep{Napoles2015GroundTF} for JFLEG evaluation. We used the CoNLL-2013 test set and JFLEG dev set as our development seed corpora for the CoNLL-2014 and JFLEG test sets, respectively.

Finally, through the process described in Section~\ref{ssec:data_synthesis}, we obtained the APR dataset illustrated in the right part of Table~\ref{table-stat}.

\subsection{Evaluation Metrics}
Since our task is to improve the readability of automatic speech transcription, the word error rate (WER), a conventional metric that is widely used in speech recognition, is not suitable for our use case. As a part of our work, instead, we investigated the usefulness and consistency of different metrics directly or modified from that of related tasks such as speech recognition, machine translation, and grammatical error correction.

\paragraph{Speech Recognition Metric} 
First, we extended the conventional WER in speech recognition to readability-aware WER (RA-WER) by removing the text normalization before calculating Levenshtein distance. We treated all mismatches due to grammatical mistakes, disfluency, as well as improper formats of capitalization, punctuation, and written numerical entity as errors. If there are alternative references, we selected the closest one to the candidate.

\paragraph{Machine Translation Metric} 
The APR task can be treated as a translation problem from a spoken transcript to a more readable written text. In this case,  we can take advantage of the BiLingual Evaluation Understudy (BLEU)~\citep{Papineni2001BleuAM} score that is widely used in machine translation to measure the performance of the APR task. In BLEU, the precision score is computed over variable-length of n-grams with length penalty \citep{Papineni2002MachineTE} and optionally with smoothing \cite{Lin2004ORANGEAM}.


\paragraph{Grammatical Error Correction Metrics}
Syntax and grammatical errors can significantly impact the readability of speech transcription. To evaluate the correctness and fluency of the rewritten sentences, we used the most commonly used GEC metrics such as MaxMatch (M$^2$) and General Language Evaluation Understanding (GLEU) in our work. M$^2$ reports the F-score of edits over the optimal phrasal alignment between the candidate and the reference sentences \citep{Dahlmeier2012BetterEF}. 
GLEU captures fluency rewrites in addition to grammatical error corrections \citep{Napoles2015GroundTF}. It is an extension of BLEU \citep{Papineni2001BleuAM} by penalizing false negatives. Besides the candidates and references used in other metrics, GEC metrics also consider source sentences in order to detect the model edits. In all experiments, we used raw ASR transcription as the source sentence when calculating GEC metrics.

\subsection{Baseline Setup}
We used the production 2-step post-processing pipeline of our speech recognition system as the APR baseline, namely n-best LM rescoring followed by inverse text normalization (ITN). This pipeline works well for sequentially improving speech recognition accuracy and display format. We computed the values of the metrics between system output of every step and the reference grammatical sentence. As a comparison, we also evaluated the original ungrammatical sentences in the same corpora. Table~\ref{baseline} shows these baseline results on CoNLL-2014 and JFLEG test sets.

\begin{table}
\centering
\resizebox{\columnwidth}{!}{%
\begin{tabular}{cl*{6}c}
\toprule[1pt]
\multirow{2}{*}{\textbf{ID}} & \multirow{2}{*}{\textbf{Candidate}} & \multicolumn{3}{c}{\textbf{CoNLL2014}} & \multicolumn{3}{c}{\textbf{JFLEG Test}}  \\
\cmidrule[0.5pt](lr){3-5} \cmidrule[0.5pt](lr){6-8}
 & & \textbf{RA-WER} & \textbf{BLEU} & \textbf{M$^2$} & \textbf{RA-WER} & \textbf{BLEU} & \textbf{GLEU} \\
\midrule[0.5pt]
a & Ungrammatical & 7.51 & 87.79 & 83.99 & 11.84 & 80.56 & 51.69 \\
\midrule[0.5pt]
\midrule[0.5pt]
b & ASR transcription & 31.15 & 60.11 & 0.00 & 29.50 & 62.42 & 20.71 \\
c & (b) + ITN & 22.01 & 70.28 & 65.76 & 19.14 & 72.15 & 38.48 \\
d & (b) + LM rescoring & 30.13 & 61.82 & 17.63 & 28.76 & 63.65 & 24.38 \\
e & (d) + ITN & \textbf{20.70} & \textbf{72.43} & \textbf{68.37} & \textbf{18.22} & \textbf{73.71} & \textbf{42.42} \\
\bottomrule[1pt]
\end{tabular}
}
\caption{\label{baseline} Baseline results are shown. Source sentences are the raw ASR transcriptions, which are the output of the TTS and ASR pipeline obtained from the ungrammatical inputs in CoNLL2014 and JFLEG test sets. References are the corresponding corrected sentences in the two corpora. Candidates are the outputs of each step of the baseline system. (a): the original ungrammatical sentence before data synthesis, which is for reference; (b): the raw ASR transcription, the same with the source; (c): ASR transcription followed by ITN; (d): ASR transcription followed by second-pass LM rescoring; (e): ASR transcription followed by LM rescoring and then ITN. }
\end{table}



An interesting finding is that although the original ungrammatical sentences in JFLEG have more errors or are less smooth than the ones in CoNLL2014 according to the higher RA-WER (11.84 vs. 7.51) and lower BLEU (80.56 vs. 87.79), the situation is inverted after transforming the sentences to and back from speech (29.50 vs 31.15 in RA-WER and 62.42 vs 60.11 in BLEU). This result may indicate that: 1) JFLEG annotators focused more on fluency and formality of the rewriting rather than pure and token-level error corrections in CoNLL2014, and 2) the ASR system, due to the powerful language model, has the ability to regularize input errors and make the transcription appear more fluent and formal. Second, GEC metrics are more sensitive to correct edits than other metrics due to the consideration of the input source sentences. Third, ITN consistently shows a much more significant impact than LM rescoring, which demonstrates the importance of display format in readability and also raises the question of how to further emphasize the correctness for future work.

\subsection{Models}

In this work, we compare different Transformer \citep{Vaswani2017AttentionIA} architectures together with corresponding open-sourced pre-trained models.

\subsubsection{MASS\footnote{{https://github.com/microsoft/MASS}}}

MASS~\citep{song2019mass} adopts the encoder-decoder framework to reconstruct a sentence fragment given the remaining part of the sentence. 
This framework is ideally suited for our task.

Following the MASS setting, we tokenized the data using the Moses toolkit\footnote{{https://github.com/moses-smt/mosesdecoder}} 
and used the same BPE codes and vocabulary from MASS.

We fine-tuned the model based on the weights pre-trained on English monolingual data. 
The model consists of a 6-layer encoder and a 6-layer decoder.
The learning rate was $10^{-4}$ with linear warm-up beginning from $10^{-7}$ for the first 4K updates, followed by inverted squared decay.
To fully utilize the GPU, we use dynamically sized mini-batches with 3000 tokens per batch.

\subsubsection{UniLM\footnote{{https://github.com/microsoft/unilm}}}

UniLM~\citep{dong2019unified} was pre-trained using the BERT-large \citep{devlin2018bert} architecture and three types of language modeling tasks: unidirectional, bidirectional, and sequence-to-sequence prediction. The unified modeling approach 
allows UniLM to be used for both discriminative and generative tasks.

Following the UniLM setting, we tokenize the training data using WordPiece \citep{Wu2016GooglesNM} with vocabulary size 28,996.
The model is a 24-layer Transformer with around 340M parameters. 

We fine-tuned the model 
for 4 epochs
. 
The learning rate was $10^{-5}$, with linear warmup over the one-tenths of total steps and linear decay. 
The batch size, maximum sequence length and masking probability were set to 256, 192 and 0.7, respectively. 
We also used label smoothing \citep{szegedy2016rethinking} with a rate of 0.1. 

Following standard practice, we removed duplicate trigrams in beam search and tuned the maximum output length and length penalty on the development set \citep{Paulus2017ADR,fan2017controllable}.

\subsubsection{RoBERTa\footnote{{https://github.com/pytorch/fairseq}}}

RoBERTa~\citep{liu2019roberta} is a robustly optimized BERT~\citep{devlin2018bert} pre-training approach
. Both BERT and RoBERTa have single Transformer stack and are pre-trained only using bidirectional prediction, which makes them more discriminative than generative. However, ~\citet{hrinchuk2019correction} demonstrated the effectiveness of transfer learning from BERT to sequence-to-sequence task by initializing both encoder and decoder with pre-trained BERT in their speech recognition correction work.

Inspired by this work and UniLM, we applied self-attention masks on the RoBERTa model to convert it into a sequence-to-sequence generation model. To achieve whole-sentence prediction rather than only masked-position prediction, we used an autoregressive approach during the fine-tuning. Another benefit from this approach is that the model can predict the end of sentence precisely; hence, there is no need to tune the maximum output length and length penalty as in UniLM fine-tuning.

Following the RoBERTa setting, the sentences were tokenized with a byte-level BPE tokenizer. The vocabulary size was 50,265. We fine-tuned the model based on RoBERTa-large pretrained weights.

\subsubsection{RoBERTa-UniLM}
Besides using UniLM and RoBERTa, we also experimented to leverage the advantages of both works to further enhance the pre-trained model before fine-tuning it on APR task. We adapted RoBERTa-large model by training it longer on a combination of English Wikipedia\footnote{{https://dumps.wikimedia.org/enwiki/latest/enwiki-latest-pages-articles.xml.bz2}}, Books\footnote{{https://www.smashwords.com/books/category/1/downloads/0/free}}, and News-Crawl\footnote{{http://data.statmt.org/news-crawl/en-doc}} data, totaling to 66GB of uncompressed text. The training was similar to UniLM but also included autoregressive (both left-to-right and right-to-left) prediction. We kept next-sentence objective in the bidirectional masked LM (MLM). All predictions conducted by whole-word masking. The first three predictions also had bigram, trigram, and phrase masking each on about 10\% of the training instances. 
We used Huggingface Transformers\footnote{{https://github.com/huggingface/transformers}} code for LM fine-tuning. The RoBERTa-UniLM model was trained for 10 days on 64 NVIDIA DGX-2 GPU cards for 7,200 steps with a batch size of 12,800. The learning rate was $10^{-4}$, with the same warmup and decay strategy with UniLM fine-tuning. The APR task fine-tuning and decoding were the same with the RoBERTa experiment.

In all fine-tuning experiments described above, checkpoints were selected on the development set, and the beam size for beam search was set to 5.

\section{Results and Discussion}

\subsection{Dataset Selection}
\label{para:data-selection}

As described in Section~\ref{ssec:data_synthesis}, we constructed the APR data using TTS and ASR systems. When an audio file synthesized by TTS is inputted to the ASR system, it will generate multiple candidate sentences from the beam search for re-ranking. These sentences have a few words that are different from the final output. At first, we used all of these sentences as our APR training data. When training our model on this data, we found that the loss converges very quickly. It usually takes only one-fourth epoch to converge. We inspected the data and found that top-K sentences produced by beam search lack in diversity, often differing only in a few characters. 
To further verify our assumption, we conducted an experiment on the MASS model and training data with different sizes.

\begin{table}
\small
\centering
\begin{tabular}{lccc}
\toprule[1pt]
\textbf{Data Size} & \textbf{RA-WER} & \textbf{BLEU} & \textbf{M$^2$} \\
\midrule[0.5pt]
LARGE (18.6M) & 18.96 & 74.90 & 71.05 \\
MODERATE (16M) & \textbf{17.15} & \textbf{76.20} & \textbf{71.76} \\
SMALL (1.1M) & 18.56 & 74.51 & 71.43 \\
\midrule[0.5pt] \midrule[0.5pt]
ORIGIN (1.1M) & 24.46 & 67.28 & 51.80 \\
\bottomrule[1pt]
\end{tabular}
\caption{\label{dataset-comporison}
Evaluation of MASS model that is fine-tuned on different size training dataset is shown. The values are evaluated on the CoNLL test set. The numbers in the parentheses are the approximate number of sentence pairs. MASS trained on \textbf{MODERATE} data achieves the best scores on all metrics. MASS trained on \textbf{SMALL} data gets a comparable result to the highest scores with a significantly smaller dataset. \textbf{ORIGIN} is the original GEC sentence pairs, which are used as the seed corpus for the APR task.  
}
\end{table}

Table~\ref{dataset-comporison} shows the results. \textbf{SMALL} data only includes the best output of the beam search and has 1.1M sentence pairs.
\textbf{MODERATE} data contains top-k beams obtained with the beam search and has 16M sentences pairs.
\textbf{LARGE} is the largest data, which also comprises original GEC pairs and TTS normalized data in addition to all data in \textbf{MODERATE}. \textbf{LARGE} has 18.6M sentence pairs. To demonstrate the difference between the GEC task and APR task, we also used the original GEC pairs as the training data denoted as \textbf{ORIGIN}.

In table~\ref{dataset-comporison}, we can see \textbf{MODERATE} data get the best scores on all metrics. That means that including top-k beams indeed helped the APR task. However, by only using \textbf{SMALL} data, we still got a comparable result. The remaining data ($\approx$15M) yielded a 1.69 increase on BLEU. This result proves our assumption that the top-k beams obtained with the beam search are homogeneous, which is not very beneficial for the model to learn new patterns from the data. Given these results and efficient usage of computational resources, we used \textbf{SMALL} data in the remainder of our experiments.
It is interesting that the \textbf{LARGE} data got a lower score than the \textbf{SMALL} data. We think the cause is the original GEC pairs, and TTS normalized data having different patterns with the ASR output data. 

The last row of Table~\ref{dataset-comporison} is the \textbf{ORIGIN} data, which has the same target sentences with the \textbf{SMALL} data but differs in source sentences. The MASS model trained on GEC pairs only got 67.28 BLEU, which is much lower than any dataset with ASR output as the source. It shows that the GEC task is different from the APR task, and APR is a new task that deserves a dedicated research effort.

\subsection{Model Comparison}

In Table~\ref{model_comparison}, we report the experimental results of four fine-tuned models on \textbf{SMALL} dataset (1.1M sentence pairs) and compared them with the baseline method. 

\begin{table}
\centering
\resizebox{\columnwidth}{!}{%
\begin{tabular}{l*{6}c}
\toprule[1pt]
\multirow{2}{*}{\textbf{Model}} & \multicolumn{3}{c}{\textbf{CoNLL2014}} & \multicolumn{3}{c}{\textbf{JFLEG}}  \\
\cmidrule[0.5pt](lr){2-4} \cmidrule[0.5pt](lr){5-7}
 & \textbf{RA-WER} & \textbf{BLEU} & \textbf{M$^2$} & \textbf{RA-WER} & \textbf{BLEU} & \textbf{GLEU} \\
\midrule[0.5pt]
Rescoring + ITN$^*$ & 20.70 & 72.43 & 68.37 & 18.22 & 73.71 & 42.42 \\
MASS & 18.56 & 74.51 & 71.43 & 18.37 & 75.37 & 47.64 \\
UniLM & 18.06 & 76.32 & 72.94 & 17.10 & 76.58 & \textbf{51.78} \\
RoBERTa & \textbf{16.59} & \textbf{77.38} & 73.65 & \textbf{13.88} & 80.34 & 51.21 \\
RoBERTa-UniLM & 16.62 & 77.24 & \textbf{74.83} & 14.13 & \textbf{80.77} & 51.16 \\
\bottomrule[1pt]
\end{tabular}
}
\caption{\label{model_comparison}
Experimental results of the baseline method and four fine-tuned models on the APR task are shown. The best values are highlighted in bold font. For RA-WER, lower is better. For other metrics, higher is better.
The results are evaluated on the CoNLL2014 and JFLEG datasets, respectively.
$^*$ is the baseline method.
}
\end{table}

Compared with the 2-step pipeline baseline, all the fine-tuned model got better scores on almost all metrics, which proved the effectiveness of considering the APR as a sequence-to-sequence task and utilizing a pre-trained model. The only exception is that MASS got a higher RA-WER (18.37) than the baseline.

In four fine-tuned models, MASS had a lower performance compared to the other three. This is reasonable since MASS only has a 6-layer encoder and 6-layer decoder, which is equivalent to a 12-layer BERT base model with about 110M parameters, while the other three are all based on 24-layer BERT-large model with about 340M parameters. The result proved again that high capacity Transformer architecture had a positive impact on the ASR task. To compare the experimental results fairly, we will focus on three BERT-large based model in the following discussion.

RoBERTa and RoBERTa-UniLM model achieved better scores than UniLM in all metrics except GLEU on JFLEG test sets. Our experiments demonstrated that fine-tuned downstream tasks based on RoBERTa gave better performance than based on BERT, which is consistent with the RoBERTa paper \citep{liu2019roberta}.

For CoNLL2014 test set, the three BERT-large based models got the comparable scores. While RoBERTa won in RA-WER and BLEU metrics, RoBERTa-UniLM won in M$^2$. CoNLL2014 test set includes minimal edits which correct the grammatical errors of a sentence but do not necessarily make it fluent or native-sounding. 
For JFLEG test set, an interesting finding is that although RoBERTa based model has fewer errors or more smooth than UniLM according to the lower RA-WER (13.88 vs. 17.10) and higher BLEU (80.77 vs. 76.58), the situation is opposite when considering fluency which is evaluated by GLEU (51.78 vs. 51.21). Unlike CoNLL, JFLEG does not restrict corrections to minimal error spans. Instead, it contains holistic sentence rewrites, which represent a diversity of potential corrections with four corrections of each sentence. Therefore, the less error and better alignment are not equivalent to the fluency or sound natural to a native English speaker, which means the metrics from ASR/MT/GEC may not be the best metrics for APR, and we need to find more suitable metrics for APR which considers more readability.

RoBERTa-UniLM did not show significant benefits than RoBERTa overall. One possible reason could be fewer LM training data. RoBERTa used 160GB text while RoBERTa-UniLM used 66GB. Another reason could be the training time. Due to the limitation of computation resources, we only updated the model for 7200 steps, and the training has not fully converged yet. However, even with limited training data and model updates, RoBERTa-UniLM demonstrated its potential in some metrics such as M$^2$ on CoNLL2014 and BLUE on JFLEG, while other metrics had very close scores to RoBERTa. This encouraged us to continue to improve the pre-trained model in the future.

In table~\ref{table-examples}, we show some examples generated by RoBERTa-UniLM. In these samples, the model correctly added punctuation, capitalization, changed the number formatting and corrected the recognition errors.


\begin{table}
\small
\centering
\begin{tabular}{p{0.95\columnwidth}}
\toprule[1pt]
\textbf{Input:} \textit{when we're diagram honest out} with certain genetic \textit{disease} are we supposed to disclose this result to our \textit{relatives} \\
\textbf{Output:}\textit{When we are diagnosed} with \textit{a} certain genetic \textit{disease}, are we supposed to disclose this result to our \textit{relatives}? \\
\midrule[0.5pt]
\textbf{Input:} \textit{this} has been indicated in report done by \textit{boston university} in \textit{two thousand and twelve eighty percent} of information posted online is \textit{true other} person may \textit{like} three out of ten times in real \textit{life} \\
\textbf{Output:} \textit{This} has been indicated in \textit{a} report done by \textit{Boston University} in \textit{2012}. \textit{80\%} of information posted online is \textit{true, although a} person may \textit{lie} three out of ten times in real \textit{life}.\\
\midrule[0.5pt]
\textbf{Input:} \textit{on} the other \textit{hand} although in \textit{turkey} the public transportation is getting \textit{developed} especially in the two big \textit{cities that is an anchor in istanbul} it is still nothing compared to \textit{sweden}\\
\textbf{Output:} \textit{On} the other \textit{hand,} although in \textit{Turkey} the public transportation is getting \textit{developed,} especially in the two big \textit{cities, Ankara and Istanbul,} it is still nothing compared to \textit{Sweden.}\\
\midrule[0.5pt]
\textbf{Input:} \textit{we} have a lot of \textit{paper have} to \textit{write}\\
\textbf{Output:} \textit{We} have a lot of \textit{papers} to \textit{write.}\\
\midrule[0.5pt]
\textbf{Input:} \textit{in} my \textit{SC i} want to focus on how important is for students to learn \textit{facts}\\
\textbf{Output:} \textit{In} my \textit{essay, I} want to focus on how important \textit{it is} for students to learn \textit{facts.} \\ 
\bottomrule[1pt]
\end{tabular}
\caption{\label{table-examples} 
Samples generated by the RoBERTa-UniLM model are shown. The italic font indicates the places of modifications.}
\vspace{-10pt}
\end{table}



\vspace{-10pt}
\section{Conclusion}
In this work, we proposed a new NLP task named ASR 
Post-processing for Readability (APR) that aims to correct grammar mistakes, disfluency, and readibility of the ASR output. We described our process for synthesizing a dataset for the APR task using the GEC datasets as our seed corpus by using TTS and ASR systems. We borrowed metrics from similar tasks and extended WER into readability-aware WER. We experimented with different dataset sizes and compared different models (MASS, UniLM, RoBERTa, RoBERTa-UniLM) with a traditional post-processing system. The results show that the fine-tuned models improved the readability of ASR output significantly, hinting at potential benefits of the APR task. We hope that our findings will encourage other researchers to work on improving readability in speech transcription systems. APR is an interesting research topic that can be considered as a style transfer from informal spoken language to a written, more formal language.


\bibliography{acl2020}
\bibliographystyle{acl_natbib}

\end{document}